\documentclass[sigconf]{acmart}

\AtBeginDocument{%
  }

\usepackage[utf8]{inputenc}
\usepackage{soul}
\usepackage{nicefrac}
\usepackage{multirow}
\usepackage{algorithm}
\usepackage{algpseudocode}
\usepackage{enumitem}
\usepackage{array}
\usepackage{xspace}

\newcommand{\uparrowcell}{\(\uparrow\)}
\newcommand{\downarrowcell}{\(\downarrow\)}
\algnewcommand\algorithmicinput{\textbf{Input:}}
\algnewcommand\Input{\item[\algorithmicinput]}
\algnewcommand\algorithmicoutput{\textbf{Output:}}
\algnewcommand\Output{\item[\algorithmicoutput]}

\renewcommand{\shortauthors}{Jian Chen et al.}
\copyrightyear{2025}
\acmYear{2025}
\setcopyright{acmlicensed}\acmConference[KDD '25]{Proceedings of the 31st ACM
SIGKDD Conference on Knowledge Discovery and Data Mining V.2}{August 3--7,
2025}{Toronto, ON, Canada}
\acmBooktitle{Proceedings of the 31st ACM SIGKDD Conference on Knowledge
Discovery and Data Mining V.2 (KDD '25), August 3--7, 2025, Toronto, ON, Canada}
\acmDOI{10.1145/3711896.3737406}
\acmISBN{979-8-4007-1454-2/2025/08}

\begin{document}
\title{\textit{ClimateIQA}: A New Dataset and Benchmark to Advance Vision-Language Models in Meteorology Anomalies Analysis}
\renewcommand{\shorttitle}{\textit{ClimateIQA}: A New Dataset and Benchmark}
\author{Jian Chen}
\orcid{0009-0005-6264-0630}
\affiliation{%
  \institution{Thrust of Artificial Intelligence, The Hong Kong University of Science and Technology (Guangzhou), HSBC}
  \city{Guangzhou}
  \country{China}
}
\email{jchen524@connect.hkust-gz.edu.cn}

\author{Peilin Zhou}
\orcid{0000-0001-6763-5236}
\affiliation{%
  \institution{Thrust of Data Science and Analytics, The Hong Kong University of Science and Technology (Guangzhou)}
  \city{Guangzhou}
  \country{China}}
\email{pzhou460@connect.hkust-gz.edu.cn}

\author{Yining Hua}
\orcid{0000-0001-7779-1208}
\affiliation{%
  \institution{Harvard University}
  \state{MA}
  \country{United States}
}
\email{yininghua@g.harvard.edu}

\author{Dading Chong}
\orcid{0000-0003-3495-522X}
\affiliation{%
 \institution{Peking University}
 \city{Shenzhen}
 \country{China}}
\email{1601213984@pku.edu.cn}

\author{Meng Cao}
\orcid{0000-0002-8946-4228}
\affiliation{%
  \institution{Mohamed bin Zayed University of Artificial Intelligence}
  \city{Masdar}
  \country{United Arab Emirates}}
\email{mengcaopku@gmail.com}

\author{Yaowei Li}
\orcid{0000-0003-0725-6108}
\affiliation{%
  \institution{Harvard University}
  \state{MA}
  \country{United States}}
\email{yaoweili@seas.harvard.edu}

\settopmatter{authorsperrow=4}

\author{Wei Chen}
\orcid{0009-0003-2260-9079}
\affiliation{%
  \institution{Thrust of Data Science and Analytics, The Hong Kong University of Science and Technology (Guangzhou)}
  \city{Guangzhou}
  \country{China}}
\email{onedeanxxx@gmail.com}

\author{Bing Zhu}
\orcid{0000-0002-2161-5974}
\affiliation{%
  \institution{HSBC}
  \city{Shanghai}
  \country{China}}
\email{bing1.zhu@hsbc.com}

\author{Junwei Liang}
\authornote{Corresponding Author}
\orcid{0000-0003-2219-5569}
\affiliation{%
  \institution{Thrust of Artificial Intelligence, The Hong Kong University of Science and Technology (Guangzhou)}
  \city{Guangzhou}
  \country{China}}
\email{junweiliang@hkust-gz.edu.cn}

\author{Zixuan Yuan}
\authornotemark[1]
\orcid{0000-0003-1197-0347}
\affiliation{%
  \institution{Thrust of Financial Technology, The Hong Kong University of Science and Technology (Guangzhou)}
  \city{Guangzhou}
  \country{China}}
\email{zixuanyuan@hkust-gz.edu.cn}
%% headers.
\renewcommand{\shortauthors}{Jian Chen et al.}
\begin{abstract}
Meteorological heatmaps play a vital role in deciphering extreme weather phenomena, yet their inherent complexities—marked by irregular contours, unstructured patterns, and complex color variations—present unique analytical hurdles for state-of-the-art Vision-Language Models (VLMs). Current state-of-the-art models like GPT-4o, Qwen-VL, and LLaVA 1.6 struggle with tasks such as precise color identification and spatial localization, resulting in inaccurate or incomplete interpretations. To address these challenges, we introduce Sparse Position and Outline Tracking (SPOT), a novel algorithm specifically designed to process irregularly shaped colored regions in visual data. SPOT identifies and localizes these regions by extracting their spatial coordinates, enabling structured representations of irregular shapes. Building on SPOT, we construct \textit{ClimateIQA}, a novel meteorological visual question answering (VQA) dataset, comprising 26,280 high-resolution heatmaps and 762,120 instruction samples for wind gust, total precipitation, wind chill index and heat index analysis. \textit{ClimateIQA}~enhances VLM training by incorporating spatial cues, geographic metadata, and reanalysis data, improving model accuracy in interpreting and describing extreme weather features. Furthermore, we develop Climate-Zoo, a suite of fine-tuned VLMs based on SPOT-empowered \textit{ClimateIQA}, which significantly outperforms existing models in meteorological heatmap tasks. 
\end{abstract}

%%
%% Keywords.
\maketitle

\newcommand\kddavailabilityurl{10.5281/zenodo.11635522}
\ifdefempty{\kddavailabilityurl}{}{
\begingroup\small\noindent\raggedright\textbf{KDD Availability Link:}\\
The source code of this paper has been made publicly available at \url{\kddavailabilityurl}.
\endgroup
}

\section{Introduction}
\begin{figure*}[t]
    \centering
    \includegraphics[width=1\linewidth]{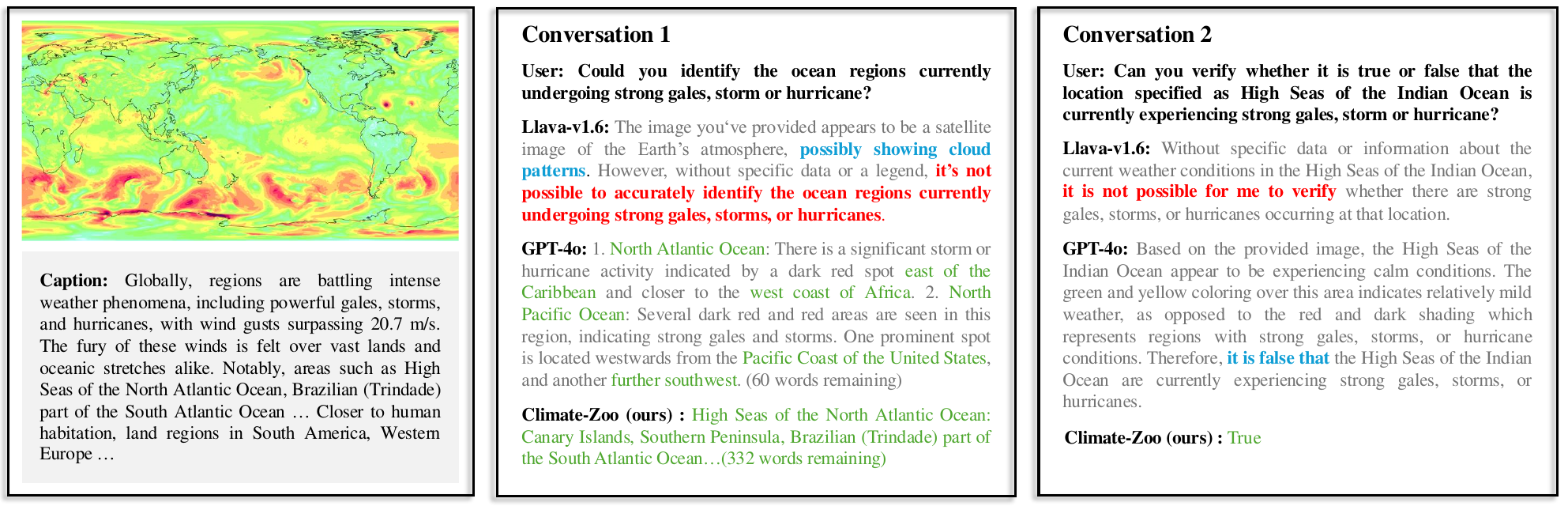}
    \caption{Comparative Analysis of Visual Chat and Reasoning Abilities in Meteorological Anomalies Analysis. Regions marked in yellow indicate strong breezes, red indicates hurricanes, and green indicates moderate breezes. In the conversation, hallucinations are marked in \textcolor{blue}{blue}, refusal-to-answer responses are marked in \textcolor{red}{red}, and accurate responses are marked in \textcolor{green}{green}.}
    \label{fig:case-study}
\end{figure*}

The focus of meteorological anomaly analysis is to identify atypical weather patterns that deviate from normal conditions, such as extreme precipitation and extreme drought \citep{seneviratne2021weather}. Accurate detection of these anomalies is crucial for improving weather forecasting and disaster preparedness \citep{fathi2022big}. Meteorological heatmaps are a key tool in this analytical process, as they visually and intuitively represent data such as wind speed, precipitation, and temperature gradients, facilitating the localization and identification of anomalous regions and types by researchers. However, manual analysis of these heatmaps is a time-consuming and error-prone task due to their often chaotic spatial distribution, complex color schemes, and irregular shapes. Although this task relies heavily on the expertise and domain knowledge of meteorological specialists \citep{wunsch2022deep}, existing automated annotation tools are limited in effectively addressing these challenges. Consequently, developing efficient and automated methods for the robust analysis of meteorological anomalies remains an underexplored area.

Recent advances in Vision-Language Models (VLMs) offer a promising solution to this challenge, as they can simultaneously process complex visual patterns and textual data~\citep{achiam2023gpt,liu2024visual}, providing potential textual interpretive labels for visual images through their inherent knowledge. Leveraging their multi-modal understanding capabilities, VLMs have achieved significant success in multiple domains, such as identifying tumors in X-ray images in the medical field and monitoring deforestation in satellite images in the ecological domain \citep{bordes2024introduction, dang2024explainable}. This naturally raises the question: \textit{Can current VLMs accurately interpret meteorological heatmaps and provide specialized textual meteorological anomaly analysis?}

To explore this, we conducted a preliminary evaluation by testing two leading VLMs, GPT-4o \citep{achiam2023gpt} and LLaVA 1.6 \citep{liu2024visual}, on tasks tailored for meteorological anomaly analysis. Specifically, as shown in Fig.~\ref{fig:case-study}, we prompted the models to enumerate anomalies (Conversation 1) and verify anomalies in specific regions (Conversation 2) within heatmaps containing extreme weather patterns. 
In Sec.\ref{sec:initial}, we further conducted four experiments to assess GPT-4o’s ability to identify and localize red regions in heatmap images, summarizing the results in Fig.~\ref{fig:preliminary-study}. 
These preliminary experiments demonstrate that the models underperformed, exhibiting several key limitations:
1) \textit{Color Misinterpretation:} VLMs frequently confuse color semantics. In  Figure \ref{fig:preliminary-study}, GPT-4o mislabels yellow regions as red within the images;
2) \textit{Hallucinations:} Models generate fictitious details absent from input data. As shown in Fig.~\ref{fig:case-study}, LLaVA 1.6 erroneously attributes strong gales to cloud patterns;
3) \textit{Incomplete Spatial Reasoning:} VLMs omit critical regions during enumeration. In Figs.~\ref{fig:case-study} and \ref{fig:preliminary-study}, GPT-4o fails to list all critical anomalies.

The poor performance of general VLMs on meteorological analysis tasks is understandable, as these models are not trained to capture the spatial details and geographic knowledge inherent in meteorological data. To address this gap, it is imperative to construct an unbiased, effective, and high-quality meteorological visual question-answering (VQA) dataset that incorporates both visual details  (e.g., matching temperature levels) and location information (e.g., corresponding GPS coordinates). Some related work has been done in this direction, but it is still insufficient. On one hand, conventional geographical representation methods (e.g., PolyWorld \citep{zorzi2022polyworld} and centroid-based representations) fail to model the irregular geometric shapes of weather systems, while segmentation models like SAM \citep{kirillov2023segment} and Lisa \citep{lai2024lisa} cannot structurally represent shapes through coordinate sequences. As a result, \textit{these methods struggle to effectively construct the chaotic spatial patterns in heatmaps.} On the other hand, some works, such as ExtremeWeather \citep{racah2017extremeweather}, focus on extreme numerical weather but ignore visual reasoning, while ClimSim \citep{yu2024climsim} prioritizes climate simulation over the interpretation of real-world heatmaps. Even meteorology-oriented VQA datasets like Terra \citep{NEURIPS2024_7a6a7fbd} lack a ground truth of geographical knowledge, thereby limiting their utility for spatial anomaly localization. Therefore, \textit{these datasets fail to effectively integrate visual and textual domain-specific knowledge.}

To address these gaps, we present \textit{ClimateIQA}, a novel dataset explicitly designed for enhancing VLM performance in meteorological anomaly analysis. Developing \textit{ClimateIQA}~required overcoming two central challenges: (1) accurately localizing irregular meteorological features within complex visual data and (2) integrating multi-dimensional meteorological and geographic information into a unified, domain-specific framework.  
To tackle the first challenge, we propose a new algorithm called \textit{Sparse Position and Outline Tracking (SPOT)}, which extracts irregular meteorological features (e.g., extreme precipitation) using sparse coordinate points. SPOT first isolates color regions via adaptive thresholds (e.g., red = precipitation >50 mm/hr), then applies clustering to select representative points while filtering outliers, achieving precise spatial encoding with minimal data loss. For the second challenge, we incorporate ERA5 reanalysis data \citep{hersbach2020era5} alongside global sea boundaries from the IHO database \citep{flanders_marine_institute_2018} to establish real-world geographic grounding. Inspired by Chain-of-Thought reasoning (CoT), \textit{ClimateIQA}~further decomposes key anomaly analysis tasks into verification, enumeration, geo-indexing, and description question, enabling comprehensive evaluation and fine-tuning of VLMs.

Building on \textit{ClimateIQA}, we introduce \textit{Climate-Zoo}, a suite of fine-tuned VLMs specifically adapted for meteorological analysis. By leveraging leading VLM architectures such as Qwen-VL-Chat \citep{bai2023qwen}, LLaVA 1.6 \citep{liu2024visual}, and Yi-VL \citep{young2024yi}, Climate-Zoo achieves state-of-the-art performance across tasks including anomaly detection, spatial reasoning, and fine-grained heatmap interpretation. Through domain-specific fine-tuning, our models outperform general-purpose VLMs, establishing new benchmarks in meteorological applications.  
Furthermore, we explore efficient fine-tuning strategies for these large models to pave the way for broader applicability.

In summary, our key contributions are as follows:  

\begin{itemize}[left=0pt]
    \item \textbf{Sparse Position and Outline Tracking (SPOT)}: A novel algorithm for precisely representing irregular features in meteorological heatmaps using sparse, coordinate-based localization.  
    \item \textbf{\textit{ClimateIQA}~Dataset}: A comprehensive dataset comprising 26,280 annotated heatmaps and 762,120 samples, tailored for real-world meteorological anomaly analysis and visual reasoning.  
    \item \textbf{Climate-Zoo}: A family of fine-tuned VLMs that set new state-of-the-art performance benchmarks in meteorological heatmap analysis, advancing VLM capabilities in this specialized domain.
\end{itemize}

\section{Related work}
% \subsection{Irregular Polygon Shape Description}
%这个领域常被应用在GIS系统去 visualize geographic analysis。because 。过往的研究尝试使用 regular 的几何图形代表不规则几何图形。也有像PolyWorld从轮廓获取点坐标代表整个几何图形。
\subsection{AI for meteorology}
The integration of AI in meteorology has seen many applications, such as employing AI for long-term weather prediction \cite{lam2022graphcast,chen2025eac}, typhoon trajectory forecasting \cite{bi2022pangu}, and weather classification \cite{dalal2023optimized}. Models like Pangu-weather \cite{bi2023accurate}, Fengwu \cite{chen2023fengwu}, and NeuralGCM \cite{kochkov2024neural} are outstanding. The advent of LLMs like ClimSight \cite{koldunov2024local}, ChatClimate \cite{vaghefi2023chatclimate}, Arabic Mini-ClimateGPT \cite{mullappilly2023arabic}, and ClimateGPT \cite{thulke2024climategpt} has broadened the scope of textual data processing in meteorology. These models have been instrumental in assimilating general meteorological knowledge related to climates, answering common queries, and offering insights. However, these models predominantly rely on textual data. This becomes particularly limiting when addressing complex challenges such as the analysis of anomalies distributions in heatmap, where textual data alone proves inadequate and prone to inaccuracies, often leading to serious hallucinations \cite{bulian2023assessing}. Meteorologists often need to interpret data from satellite images \cite{liu2024study}, radar \cite{guastavino2022prediction}, heatmaps \cite{lee2024enhancing}, and isobaric maps \cite{xu2024increasing} to make accurate assessments. Nonetheless, there remains a lack of VLMs capable of interpreting such visual meteorological data. 
Many current AI applications in meteorology work directly with numerical or gridded data. Our work explores the domain of visual heatmaps, which presents distinct challenges and opportunities for VLMs.

\subsection{Vision language models and visual question answering}
The integration of visual and textual data has led to the development of advanced VLMs, which typically build upon the capabilities of text-only LLMs, such as GPT-4 \cite{achiam2023gpt}, LLaMA \cite{touvron2023llama}, Gemini \cite{team2023gemini}, and Claude \cite{anthropic2024claude}. Notable developments in VLMs include GPT-4o \cite{achiam2023gpt}, Qwen-VL \cite{bai2023qwen}, and LLaVA \cite{li2024llava}, which have substantially enhanced the efficiency of VQA tasks. These tasks require models to comprehend and respond to information and questions in both visual and textual formats. 

To enhance model performance in VQA, researchers have adopted advanced methods for visual feature extraction \cite{zheng2023mffn}, developed robust model architectures \cite{liu2024visual}, and explored innovative learning paradigms \cite{chen2024large}. Despite these advancements, VQA tasks continue to face challenges, such as the occurrence of hallucinations \cite{bai2024hallucination}, often stemming from issues like data quality and visual uncertainty \cite{leng2023mitigating}. Addressing these issues highlights the critical need for high-quality datasets and effective strategies to mitigate challenges in VQA tasks.

\begin{figure*}[t]
    \centering
    \includegraphics[width=1\linewidth]{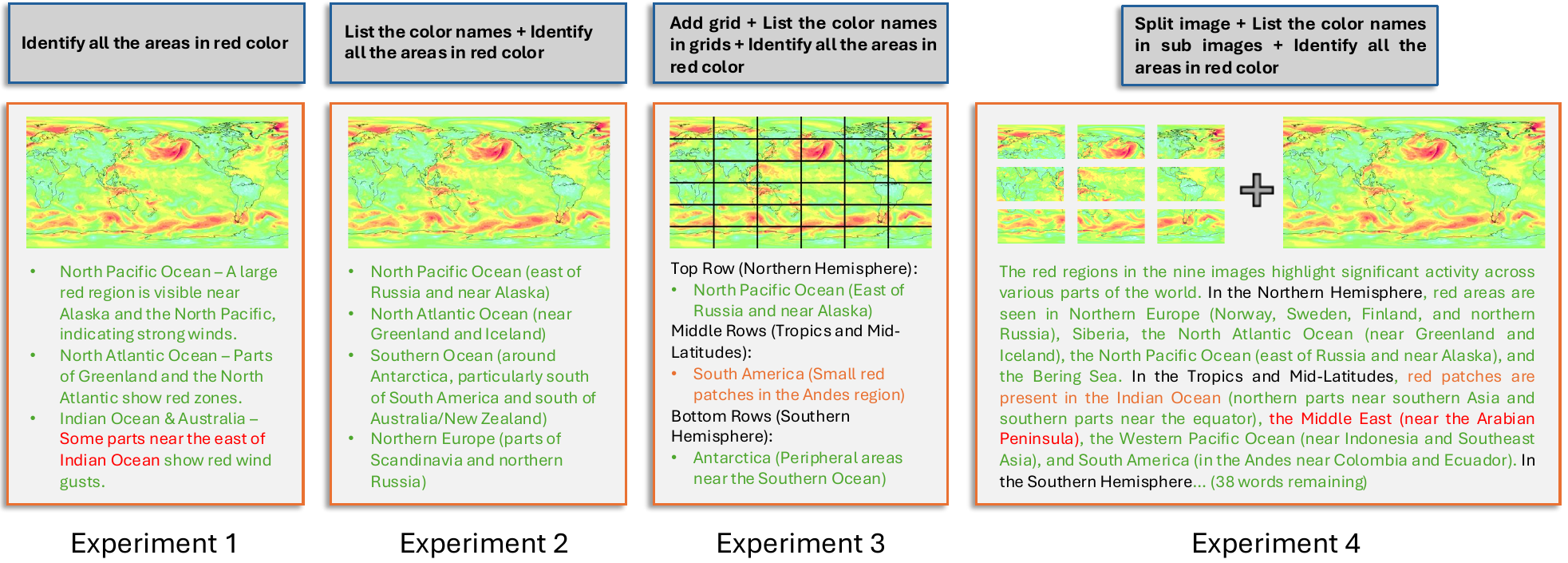}
    \caption{Result of an in-depth evaluation via Prompt-Engineering GPT-4o. Sentences in \textcolor{red}{red} mark inaccurate responses, sentences in \textcolor{orange}{orange} and black mean surprising findings (patches and geography information), and sentences in \textcolor{green}{green} mark accurate answers.}
    \label{fig:preliminary-study}
\end{figure*}

% \section{Initial assessment of GPT-4o} 
\section{Preliminary Investigation of VLM Capabilities on Meteorological Heatmaps}
\label{sec:initial}
To better understand the challenges VLMs face with meteorological heatmaps and to motivate the development of ClimateIQA, we conducted a preliminary investigation.
Among various VLMs, GPT-4o \cite{achiam2023gpt} has demonstrated exceptional capabilities in understanding and generating visual and textual content \cite{singh2024visual}. We began with an in-depth evaluation of its ability to identify and localize red regions in heatmap images, indicating areas like high wind speed, temperatures, or significant weather metrics, aiming to pinpoint areas for enhancement based on its limitations. Four experiments were designed for this assessment (Figure \ref{fig:preliminary-study}):

\begin{enumerate}[left=0pt]  % 设置列表缩进为0
    \item \textbf{Direct Red Region Identification:} We tested the VLM's ability to identify red regions directly, without guidance, to evaluate its color perception and localization capabilities.

    \item \textbf{Two-Step Color Identification:} After observing potential color confusion in the first experiment, we modified the process. The model first listed all colors in the image, then specifically identified the red regions. This approach was designed to improve the accuracy of color recognition.

    \item \textbf{Grid-Based Color Identification:} To capture fine-grained details, we divided the image into a $6 \times 6$ grid, each with geographic details. The model identified all colors present in each cell and then located the red-colored regions, allowing us to assess its ability to capture local color information and its impact on localization accuracy.

    \item \textbf{Image Segmentation and Combined Analysis:} We employed image segmentation using the PIL toolkit \cite{umesh2012image}, dividing the input image into sub-images. The VLM was tasked with analyzing both the overall and segmented images, with the results combined for a more comprehensive interpretation, aiming to improve the completeness and accuracy of the model’s responses.
\end{enumerate}

The results varied across experiments. In Experiment 1, GPT-4o struggled with direct identification of red regions, inaccurately marking locations such as "East of Indian Ocean". Experiment 2 showed improvement with correct identifications, though responses were incomplete and the recall rate was just 15\%. Experiment 3, the grid-based approach, better-captured details like patches but had inconsistent performance across different images, with an average accuracy of 18\%. Experiment 4 utilized a segmented and combined analysis approach, yielding the most accurate results among our trials. The model successfully identified sub-image colors and provided more detailed interpretations, including specific geographic coordinates and thorough annotations. Despite these improvements, the responses were still incomplete, with an average recall rate of only 22\%. Additionally, similar to Experiment 3, erroneous results occurred when segmented image analysis led to incorrect color judgments. The increased number of generated answers correlated with a higher error rate, highlighting a critical area for further enhancement.

% \section{\textit{ClimateIQA}}

\section{\textit{ClimateIQA}: Dataset Building Pipeline}

\subsection{Data collection}

\begin{figure*}[t]
    \centering
    \includegraphics[width=0.8\linewidth]{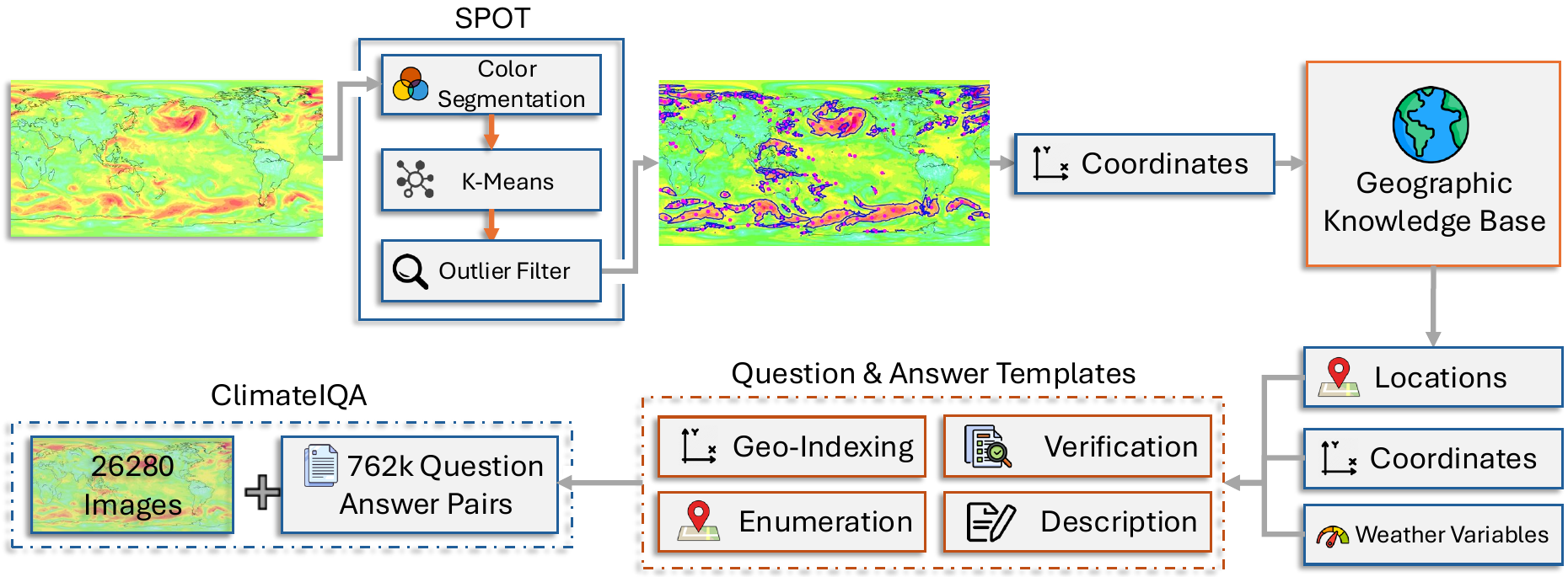}
    \caption{The process of constructing \textit{ClimateIQA}. Images were processed using SPOT to extract color contours (marked in blue) and representative point coordinates (marked in purple), such as (-40, 65). The extracted data were integrated into geographic knowledge bases to retrieve location-specific information. These data, including location, coordinates, and weather variables, were then input into predefined question-and-answer templates, resulting in the generation of 762,120 question-answer pairs. The final dataset, \textit{ClimateIQA}, pairs these QA pairs with 26,280 images, enabling comprehensive visual question answering.}
    \setlength{\belowcaptionskip}{2pt}
    \label{fig:framework}
\end{figure*}

Our meteorological data originates from the ERA5 hourly dataset on single levels, curated by the European Centre for Medium-Range Weather Forecasts (ECMWF) \cite{hersbach2020era5}. ERA5 offers a high-resolution, global repository of weather and climate data spanning back to 1940. It is constructed using advanced data assimilation techniques, where observational data is iteratively integrated with model outputs, resulting in consistent and reliable estimates that are updated every 12 hours. For this study, we focused on the year 2023 and selected specific hourly variables: wind gust, precipitation, temperature, humidity, and potential evaporation. Based on these variables, we derived three types of meteorological heatmaps: the wind speed heatmap (based on wind gust), the total precipitation heatmap (based on precipitation), and the combined wind chill and heat index heatmap (based on temperature, wind gust and humidity). 
% The samples of these heatmaps are shown in Figure \ref{fig:wg_image} - \ref{fig:temp_image} in the appendix.

To classify wind speeds, we adopted the Beaufort Scale \cite{monmonier2005defining}, a well-established categorization framework used to assess wind speeds by their physical effects on land or sea. The scale spans from 0 to 12, with each level corresponding to a specific wind speed range. For visualization purposes, we assigned a distinct color gradient to each level: beginning with white for level 0 (the calmest state), progressing through light blue, turquoise, light green, lime green, yellowish-green, light yellow, peach, light coral, salmon, deep pink, dark magenta, and culminating with dark purple for level 12 (the highest wind speeds). Meteorological literature often considers level 8 of the Beaufort Scale (20.8 m/s) as a critical threshold for extreme weather events \cite{radinovic2014measuring,weaver2021social}. Accordingly, our heatmaps highlight levels 8 and above (starting from the peach gradient) as anomalies. To facilitate spatial interpretation, the wind speed heatmap incorporates a geographical overlay of a world map that pinpoints the locations of these anomalies.

For precipitation visualization, we categorized precipitation intensity using NOAA standards \footnote{\url{https://www.noaa.gov}}, which classify precipitation into four levels: Light Precipitation, Moderate Precipitation, Heavy Precipitation, and Extreme Precipitation. Each category is visually represented by a unique color: white for light precipitation, mint green for moderate precipitation, lime green for heavy precipitation, and pale goldenrod for extreme precipitation. This classification enabled us to generate clear and interpretable precipitation heatmaps, where variations in color directly correspond to precipitation intensity.

 The combined wind chill and heat index heatmap integrates the Wind Chill Index (WCI) and Heat Index (HI) into a single visualization, capturing both cold- and heat-related scenarios across regions in 2023. To compute these indices, we followed the widely used formulas and classification standards established by NOAA:  

\begin{equation}
    \resizebox{.7\linewidth}{!}{$
\begin{aligned}
        WCI = & \  13.12 + 0.6215 \times T - 11.37 \times \left(2.23694 \times V\right)^{0.16} \\
        &+ 0.3965 \times T \times \left(2.23694 \times V\right)^{0.16}, 
\end{aligned}$}
\end{equation}%

\begin{equation}
    \resizebox{.8\linewidth}{!}{$
\begin{aligned}
    HI = & \ -42.379 + 2.04901523 \times T + 10.14333127 \times \text{hum} \\
         & - 0.22475541 \times T \times \text{hum} - 6.83783 \times 10^{-3} \times T^2 \\
         & - 5.481717 \times 10^{-2} \times \text{hum}^2 + 1.22874 \times 10^{-3} \times T^2 \times \text{hum} \\
         & + 8.5282 \times 10^{-4} \times T \times \text{hum}^2 - 1.99 \times 10^{-6} \times T^2 \times \text{hum}^2,
\end{aligned}$}
\end{equation}
 where WCI was computed for temperatures \( T \) below 4.4°C, \( T \) is the temperature in degrees Celsius and \( V \) is the wind gust speed in m/s. HI was calculated for temperatures \( T \) above 26.7°C, using the humidity percentage \( \text{(hum)} \). 
Based on the calculated indices, we grouped the results into 11 levels, ranging from extremely dangerous cold to extremely hot. Each level was assigned a specific color gradient for visualization, which transitions from dark blue (very low ranges) through tones like vivid blue, medium slate blue, and sky blue to pale turquoise for neutral conditions. Beyond neutral values, warmer conditions were represented by progressively vibrant colors: pale yellow, coral, and light red, with extreme heat visualized in shades of red. This nuanced color design enables the identification of thermal extremes, both cold and hot, within the same heatmap, thus offering a comprehensive depiction of meteorological variability.

We acknowledge that using RGB heatmaps introduces certain characteristics. These include potential projection distortions, especially at higher latitudes, and a departure from native geospatial data formats (e.g., lat/lon grids) typically used in numerical weather prediction workflows. However, this choice was made to specifically target the VLM's ability to interpret visually complex information as a human expert might and to leverage existing VLM architectures adept at processing RGB images.

\subsection{Sparse Position and Outline Tracking (SPOT)}
\begin{figure}[t]
    \centering
    \includegraphics[width=0.9\linewidth]{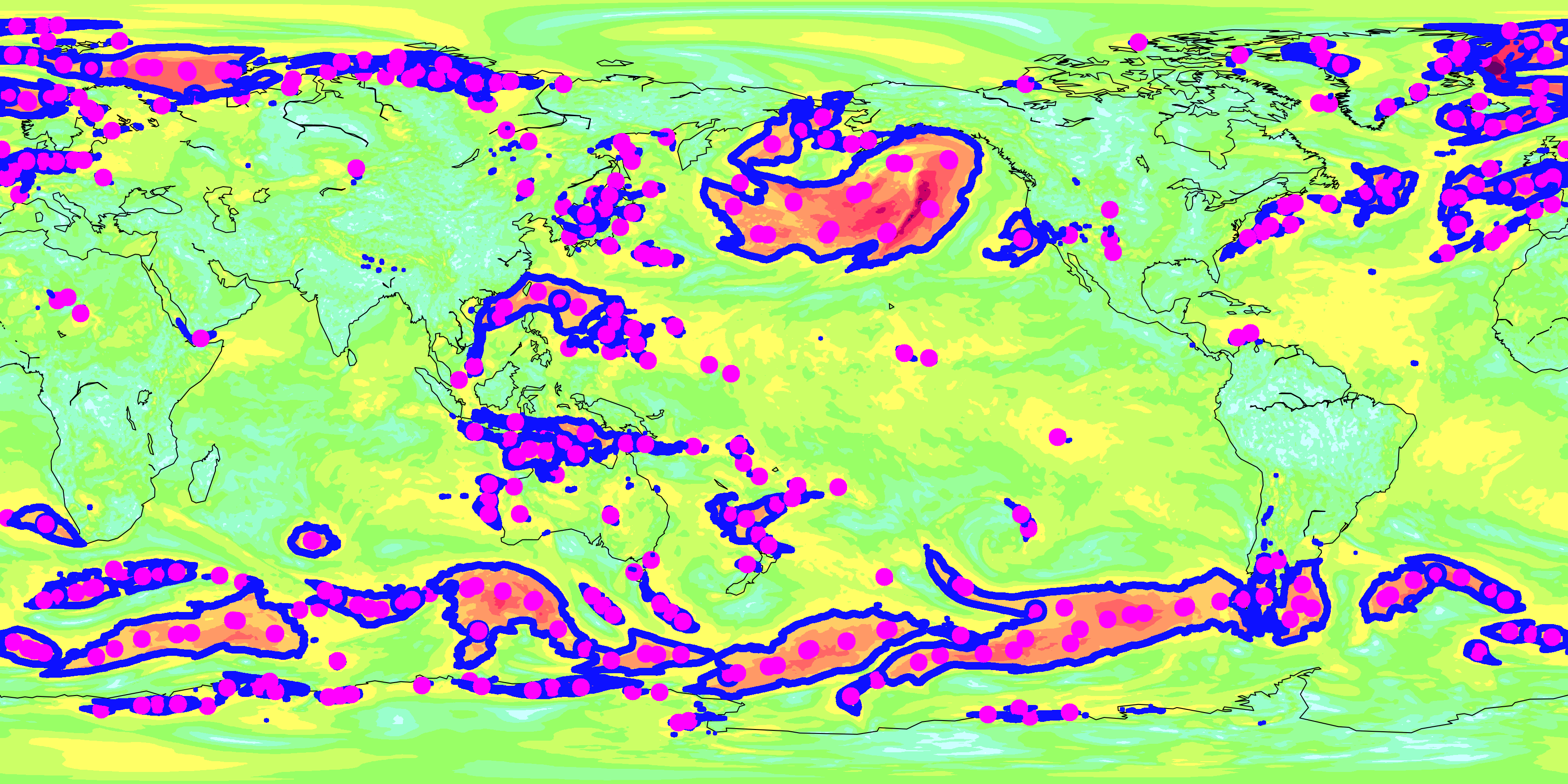}
    \caption{The SPOT algorithm identifies representative points (enlarged purple dots) within strong gale zones (light coral) from a high-resolution image, with deep blue outlines precisely tracing the contours, showing alignment of points with the contours.}
    \label{fig:spot_result}
    \vspace{-2mm}
\end{figure}

\begin{figure}[t]
    \centering
    \includegraphics[width=0.9\linewidth]{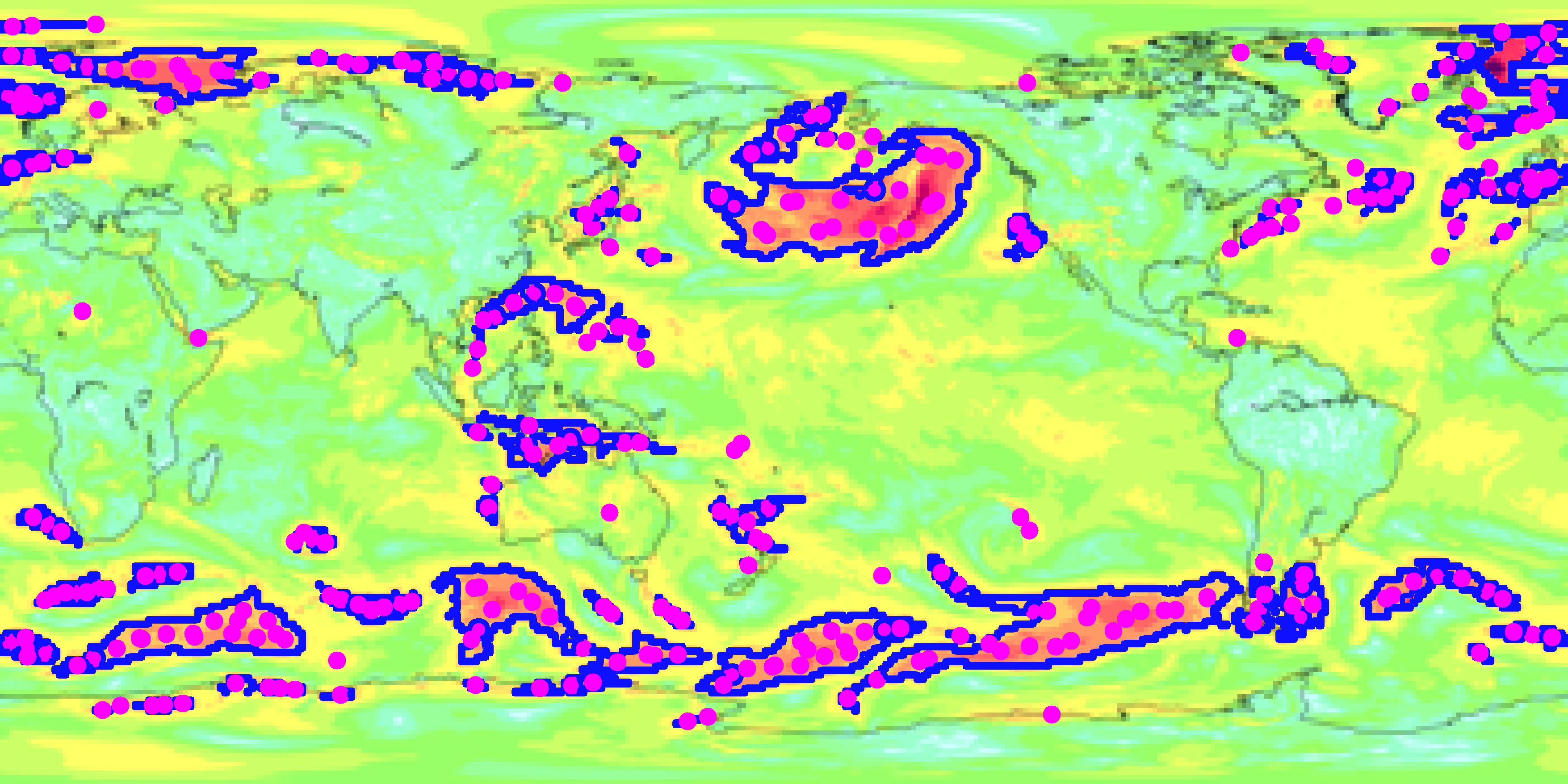}
    \caption{Low-resolution results of the SPOT algorithm. SPOT accurately outlines the shapes of light coral areas within low-resolution image, but selects fewer representative points compared to those from high-resolution heatmaps.}
    \label{fig:spot_result_lr}
    \vspace{-2mm}
\end{figure}

To address the inherent challenges of extracting meaningful patterns from irregular visual structures in meteorological heatmaps, we developed the SPOT algorithm. SPOT effectively condenses large-scale spatial data into a compact and representative format while preserving its accuracy. This reduction in data complexity accelerates model training and minimizes computational overhead. Below, we present the core stages of SPOT:

{\bf Stage 1: Color Segmentation.} Initially, our SPOT method extracts contours from heatmaps based on four primary colors: red, yellow, white, and green, using OpenCV~\cite{culjak2012brief} as the color filter . We obtain the contour coordinates of each color region to address the issue of irregular shapes often encountered in heatmaps. This process is iterated twice to ensure accuracy, selecting the best segmentation result to mitigate errors.

{\bf Stage 2. Representative Point Selection.} 
To efficiently represent the geographical location and distribution shape of each color region in the contour data, we reduce the large volume of coordinate data into a minimal set of representative points. First, we determine the number of points required based on the area of each color region within the image. Specifically, the number of points (\(k\)) is assigned as follows: 
For regions occupying less than 1\% of the total area, we assign 1 point. 
1\%-5\% of the total area: 3 points.
5\%-10\% of the total area: 5 points.
More than 10\% of the total area: 10 points.

Next, to handle the challenge of representing irregularly shaped regions, we apply the K-Means clustering algorithm to compute the centroid coordinates of each cluster within the selected region. To ensure reproducibility, we set the random state to 0. Finally, the first representative point for each region is assigned to the centroid of the initial cluster (cluster index \(k=1\)), while additional representative points maintain a direct correspondence to the spatial distribution revealed by the heatmap.

{\bf Stage 3. Filtering Outliers.} We implemented a rule-based function to ensure all points fall within their respective color regions. Any points found outside these regions are automatically excluded and replaced with new points from the nearest valid contour. In a processed heatmap containing 5,448 points, approximately 122 points may fall outside the contour, resulting in an efficiency rate of about 97.7\%. Our method reroutes these outlier points to maintain the robustness and accuracy of the model. 
As shown in Figures \ref{fig:spot_result} and \ref{fig:spot_result_lr}, we use purple dots to represent the coordinate points after applying SPOT. These dots clearly illustrate the spatial location and shape of the corresponding color regions. After color segmentation in the SPOT algorithm, representative points are assigned to their segmented color regions with 100\% accuracy. The ultimate accuracy of feature localization is dependent on the initial color segmentation quality. The pseudo-code of SPOT is detailed in Appendix \ref{alg:spot}. 

After identifying the representative points for each color block using SPOT, we indexed the corresponding geographical names of these points coordinates using two geographic databases: the IHO Sea Areas \cite{flanders_marine_institute_2018} and the World Bank-approved Administrative Boundaries \cite{world_food_programme_(un_agency)_2019}. The IHO Sea Areas database delineates the boundaries of the world's major oceans and seas, while the World Bank-approved administrative boundaries database includes international borders, disputed areas, coastlines, lakes, and a usage guide. These indexed geographical names were then used to substantiate the question-answer generation templates introduced in Section~\ref{Instruction}, which formed the basis for constructing the instruction-tuning data.

\subsection{Instruction-tuning data construction}
\label{Instruction}
The construction of high-quality instruction-tuning data is essential for enhancing the performance of VLMs in meteorological anomalies analysis. To ensure accurate and contextually relevant question-answer generation, 
we designed templates that were grounded in specific geographic data (processed by SPOT), such as location coordinates and names. 
These templates, which were reviewed and validated by human experts, provided a systematic framework for generating the instruction-tuning data.
Figure \ref{fig:conversation} shows an example of such instruction-tuning data.
Notably, as identified in Section~\ref{sec:initial}, our initial assessment of the VLM revealed several limitations, including insufficient geographic and meteorological knowledge, which led to issues such as incorrect answers, inaccurate color localization, and incomplete responses.
To address these limitations, we developed the following four question types, each targeting a specific area of improvement:
\begin{enumerate}[left=0pt,label=\textbullet] % This removes indentation
    \item \textbf{Verification Questions:} These questions determine whether a specific location in the heatmap contains anomalies. This type is designed to enhance the model’s accuracy in identifying anomalies, which is critical for timely and precise weather forecasting.
    
    \item \textbf{Enumeration Questions:} These questions list all locations in the heatmap that exhibit anomalies. The purpose of this question type is to improve the completeness of the model's responses, ensuring that all relevant aspects of a query are adequately addressed.
    
    \item \textbf{Geo-Indexing Questions:} These questions provide the coordinates of anomalies in the heatmap. Geo-Indexing questions focus on enhancing the model’s ability to accurately locate anomalies in images, which is essential for proper geographical referencing and the interpretation of meteorological data.
    
    \item \textbf{Description Questions:} These questions provide detailed interpretations of the anomalies present in the image. Description questions are intended to generate comprehensive reports, which are crucial for detailed meteorological analysis and the communication of weather-related findings.
\end{enumerate}

\begin{figure}[t]
    \centering
    \includegraphics[width=1\linewidth]{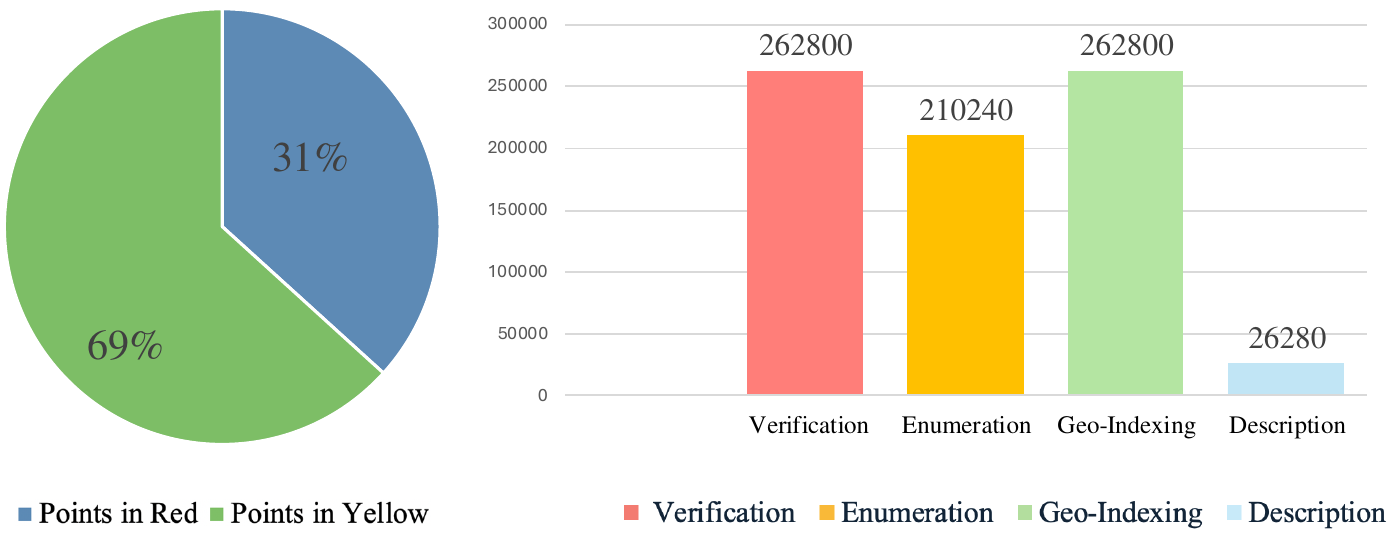}
    \caption{Distribution of red and yellow point coordinate data collected by SPOT (left) and the distribution of four question types in the \textit{ClimateIQA}~dataset (right).}
    \label{fig:chart}
    \vspace{-4mm}
\end{figure}

\vspace{-2mm}
\subsection{Dataset statistics} 

The constructed \textit{ClimateIQA}~dataset contains 26,280 high resolution heatmaps, each measuring 3510 $\times$ 1755 pixels, and a total of 762,120 instruction-tuning data points. These heatmaps provide detailed visual representations of global wind speed, total precipitation, and combined wind chill index and heat index. As shown in the right part of Figure~\ref{fig:chart}, in the instruction-tuning data, the distribution of the four question types is as follows: Verification Questions (34.5\%), Enumeration Questions (27.6\%), Geo-Indexing Questions (34.5\%), and Description Questions (3.4\%). Wind speeds exceeding 10.8 m/s, total precipitation exceeding 7.6 mm/hr, and HI exceeding 41°C are considered anomalies, which are represented in the heatmaps as red and yellow points. As shown in left part of Figure~\ref{fig:chart}, 31\% of the points collected by SPOT are red, while 69\% are yellow. The dataset was further split into training, validation, and testing sets in a 7:1:2 ratio, ensuring chronological order. This approach mimics real-world scenarios, where models are trained on historical data and evaluated on future data, thereby enhancing their ability to generalize and perform effectively on unseen instances.

\section{Climate-Zoo: Adapting VLMs to meteorology} 

This section outlines our approach to enhancing the performance of VLMs in meteorological anomaly analysis through prompt engineering and supervised fine-tuning using the \textit{ClimateIQA}~dataset. Due to the computational limitations associated with processing all heatmap types, which would require significant computational resources, we focused our experiments on wind gust heatmaps. These heatmaps are particularly suitable for anomaly analysis, as they have a higher density of color categories compared to precipitation or combined wind chill index and heat index heatmaps, making them ideal for identifying meteorological anomalies. While the current work focuses on wind gust heatmaps, we plan to extend the analysis to other types of heatmaps in future research.

\begin{figure}
    \centering
    \includegraphics[width=1\linewidth]{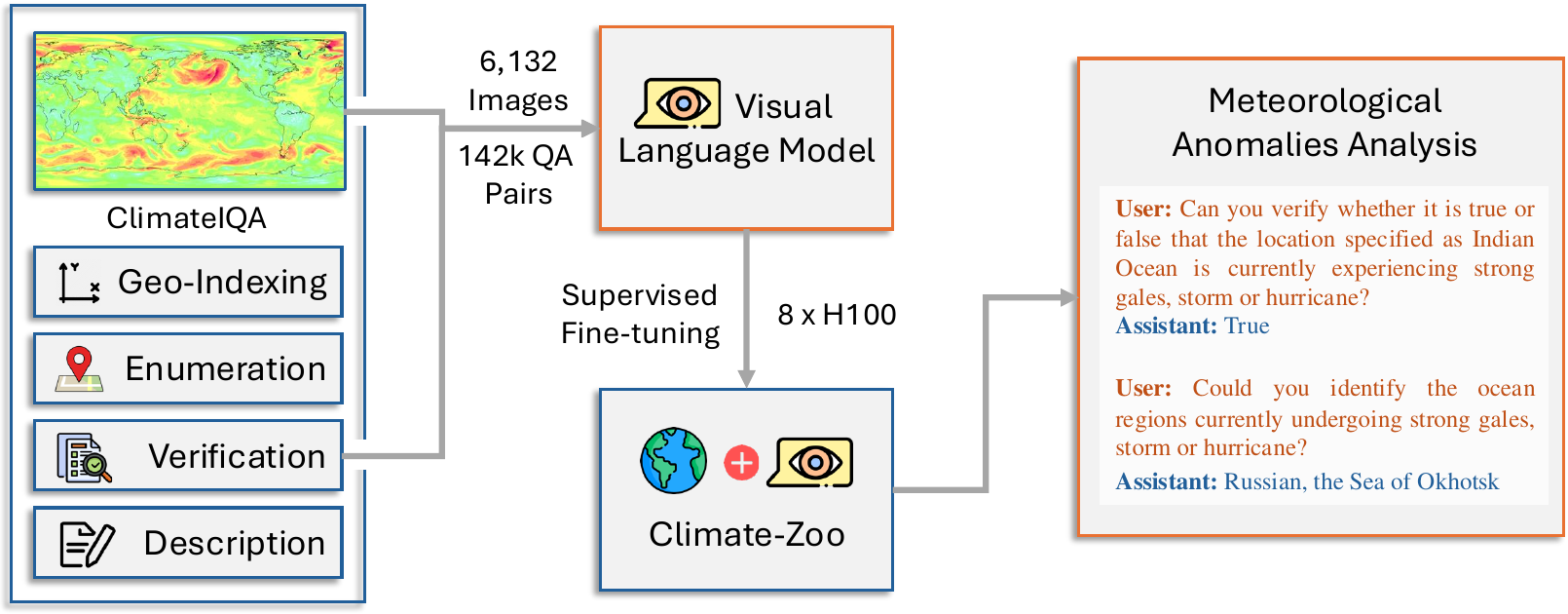}
    
    \caption{Process of adapting VLMs to meteorological anomaly analysis.}
    \label{fig:finetuning}
    \vspace{-4mm}
\end{figure}

\noindent\textbf{Base Models.}\quad
Based on model performance on VLM benchmarks \cite{goyal2017making, lu2022learn}, we selected three SOTA VLMs as our base models: LLaVA-v1.6-mistral-7b \cite{liu2024visual}, Qwen-VL-Chat \cite{bai2023qwen}, and Yi-VL-6B \cite{young2024yi}. LLaVA-v1.6 excels in multimodal understanding, Qwen-VL-Chat in visual dialog tasks, and Yi-VL-6B in visual reasoning.

\noindent\textbf{Supervised Fine-tuning (SFT).}\quad

As shown in Figure \ref{fig:finetuning}, we conducted SFT on base models using the \textit{ClimateIQA}~training set, which consists of 6,132 heatmaps and 142k QA pairs. During the SFT process, we froze the weights of the visual encoder and employed a unified encoder layer to reduce computational costs and mitigate overfitting risks. This strategy is supported by \cite{khattak2023self}, who demonstrated that pre-trained visual encoders are highly effective at extracting meaningful features. Additionally, the size of the training dataset plays a crucial role in fine-tuning performance. We experimented with different dataset sizes (10k, 50k, 100k, and 142k) to evaluate their impact on SFT effectiveness and selected the best-performing model based on its performance on the validation set.

\noindent\textbf{Training details.}\quad
We conducted supervised fine-tuning on three prominent large-scale VLMs: LLaVA-v1.6-7B, Qwen-VL-Chat-7B, and Yi-VL-6B using two distinct strategies. One strategy involved full-parameter training\footnote{The Swift toolkit was adopted for improving training efficiency and flexibility.}, while the other employed Low-Rank Adaptation (LoRA) layers for fine-tuning. LoRA introduces a low-rank decomposition of the model's weight matrices, enabling efficient adaptation to new tasks with minimal additional parameters. For LoRA fine-tuning, we set the LoRA rank to 8, the alpha value to 32, and the dropout probability to 0.05, balancing adaptation capacity and computational efficiency.

To expedite the training process, we employed 8 H100 80G GPUs and utilized Distributed Data Parallel along with DeepSpeed. The batch size was set to 1, and the learning rate was 1e-4. The entire experiment was conducted for a single epoch, spanning a total duration of 30 days.

\section{Evaluation}

\subsection{Metrics}
For each question type, we adopted different evaluation metrics tailored to its specific characteristics. We detail the evaluation metrics as follows:

 \noindent\textbf{F1 Score.}\quad
For Verification Questions, we evaluate the model's ability to judge the correctness of statements using F1 score, namely the harmonic mean of precision and recall.

\noindent\textbf{Element Match Score.}\quad 
For Enumeration Questions, we compute a match score (MS) between the ground truth ($x$) and model-generated answer ($y$). This involves comparing the sets ($x$) and ($y$) formatted as ["New York", "High Seas of the North Atlantic Ocean", "Canary Islands"], representing ground truth and model output, respectively. The match score calculation involves determining correct matches via the set intersection size (common elements in both sets ($x$) and ($y$)) and incorrect matches via the symmetric differences (elements present in one set but not in the other). In cases where both sets ($x$) and ($y$) are empty (union size of zero), the match score is defined to handle division by zero and set to zero. Otherwise, the match score ranges between -1 and 1, where a score closer to 1 indicates more accurate and complete answers with fewer hallucinations (incorrect items), and a score closer to -1 indicates poor performance with many hallucinations. The score is determined by the formula:

\begin{equation}
    \resizebox{.6\linewidth}{!}{$
    MS = 
    \begin{cases}
      0, & \text{if } |x \cup y| = 0 \\
      \frac{|x \cap y| - (|x - y| + |y - x|)}{|x \cup y|}, & \text{otherwise}
    \end{cases}
        $}
\end{equation}%

\noindent\textbf{Haversine Distance.}\quad
For Geo-indexing Questions, which involve determining precise geographical coordinates, we utilized the Haversine distance formula. This metric accurately measures the distance between model-generated coordinates ($lat_m$, $lon_m$) and ground truth coordinates ($lat_g$, $lon_g$) by accounting for the Earth's curvature. The formula is as follows, where $r$ represents the Earth's radius:

\begin{equation}
    \resizebox{.9\linewidth}{!}{$
        d = 2r \cdot arcsin(\sqrt{sin^2(\frac{lat_m-lat_g}{2}) + cos(lat_m) \cdot cos(lat_g) \cdot sin^2(\frac{lon_m-lon_g}{2})})
        $}
\end{equation}%
\noindent\textbf{BLEU, ROUGE and GPT-4o Scores.}\quad
For Description Questions, we employ average BLEU \cite{papineni2002bleu} and ROUGE \cite{lin2004rouge} and GPT-4o scores \cite{cao2024rap}. BLEU-1 and BLEU-2 measure linguistic accuracy by comparing n-grams between the generated and ground truth descriptions. ROUGE-1, ROUGE-2, and ROUGE-L assess the overlap of n-grams, word sequences, and word pairs, offering insights into the comprehensiveness and relevance of the generated descriptions. Additionally, GPT-4o evaluates the overall quality and similarity of model-generated descriptions to ground truth answers on a five-point Likert scale. 
%The prompt is shown in Appendix \ref{tab:gpt4}.

%需要加上公式吗？

\subsection{Results and analysis}

% Requires: \usepackage{graphicx}
\noindent\textbf{SPOT Across Varying Resolutions.\quad}
Table \ref{tab:resolution_data} and Figures \ref{fig:spot_result} - \ref{fig:spot_result_lr} demonstrate the performance of SPOT across images with varying resolutions. As resolution decreases, we observe that the number of selected contours and representative points declines, and the Haversine Distance increases. When compared to the ground truth points, the representative points in lower resolution images exhibit minor deviations. This discrepancy arises because SPOT struggles to detect color variations in very small regions as resolution diminishes, leading to the omission of representative points in these areas. However, the results show that SPOT effectively identifies and outlines large areas affected by extreme weather. The distribution of representative points in these regions remains accurate despite the lower resolution.

\begin{table}[t]

    \caption{SPOT results across images with varying resolutions. Contour ratio and point ratio: the number of lower-resolution detections divided by the number of ground-truth detections. }
    \label{tab:resolution_data}
    \centering
    \resizebox{0.45\textwidth}{!}{%
    \begin{tabular}{cccc}
        \hline
        \textbf{Resolution} 
        &  \textbf{\begin{tabular}[c]{@{}c@{}}Contour Ratio \uparrowcell\end{tabular}} 
        &  \textbf{\begin{tabular}[c]{@{}c@{}}Point Ratio \uparrowcell\end{tabular}} 
        &  \textbf{\begin{tabular}[c]{@{}c@{}}Haversine \\Distance \(10^3\) \downarrowcell\end{tabular}} \\
        \hline
        351 $\times$ 175.5 & 22.2\% &74.9\% & 94.205 \\
        702 $\times$ 351 & 38.3\% &88.6\% &71.712 \\
        1755 $\times$ 877.5 & 59.5\% & 94.7\%&21.566 \\
        3510 $\times$ 1755 &  100\% &100\% & 0\\
        \hline
    \end{tabular}}
\vspace{-5mm}
\end{table}

\begin{table*}[t]
\captionsetup{skip=2pt} 
\caption{Result of supervised fine-tuning}
\label{tab:sft}
\centering
\renewcommand{\arraystretch}{1} 
\resizebox{0.8\textwidth}{!}{%
\begin{tabular}{llcccccccc}\\ \hline
\multicolumn{2}{c}{\textbf{Model}}  
& \textbf{F1 Score \uparrowcell} & \textbf{\begin{tabular}[c]{@{}c@{}}Element \\ Match Score \uparrowcell\end{tabular}}
&  \textbf{\begin{tabular}[c]{@{}c@{}}Haversine \\ Distance \(10^3\) \downarrowcell\end{tabular}} & \textbf{BLEU \uparrowcell} & \textbf{ROUGE \uparrowcell} & \textbf{\begin{tabular}[c]{@{}c@{}}GPT4o Score  \\ (Similarity) \uparrowcell\end{tabular}} & \textbf{\begin{tabular}[c]{@{}c@{}}GPT4o Score \\ (Total) \uparrowcell\end{tabular}}  \\ \hline
\multirow{5}{*}{Baseline Model} & Qwen-VL-Chat &0    &-1    &69.28   &0  &0.08      &1.000        & 1.537         \\
                                & Yi-VL-6B &0    &-1    &67.18   &0.004  &0.052  &2.152  &2.983  \\
                                & LLaVA-v1.6-mistral-7b &0    &-1    &85.66   &0  &0.041    &1.744  &3.137  \\
                                %& MiniCPM-Llama3-v2.5 &0&-1&-&0 &0 &2.035 &3.199 \\
                                & GPT-4-Vision &0&-1&-&0 & 0&2.142 &3.166  \\
                                & GPT-4o &0.151&-0.684&41.56&0.324 &0.271 &2.327 &3.466  \\\hline
\multirow{3}{*}{Climate-Zoo LoRa} & Qwen-VL-Chat LoRa &0.909    &-0.930    &{1.894}    &0.819    &0.732    &\textbf{4.868}    &4.356    \\
                                & Yi-VL-6B LoRa &0.905    &-0.934    &\bf 1.887    &0.007   &0.055    &1.799    &2.902    \\
                                & LLaVA-v1.6-mistral-7b LoRa &0.910    &-0.822    &1.905    &{0.821}    &0.731    &4.658    &4.417    \\
                                %& MiniCPM-Llama3-v2.5 LoRa &0&-1&3.676 &0 &0.048 &2.019 &3.175 \\
                                \hline
\multirow{5}{*}{Climate-Zoo}    & Qwen-VL-Chat &0.910 &\textbf{-0.012} &1.928 &0.818 &0.722 &4.734 &\textbf{4.522} \\
                                & Yi-VL-6B &\textbf{0.912} &{-0.122} &1.933 &0.815 &0.728 &4.685 &4.373 \\
                                & LLaVA-v1.6-mistral-7b &0.897 &-0.483 &1.935 &\bf 0.823 &\textbf{0.747} &4.798 &4.423 \\
                                %& MiniCPM-Llama3-v2.5 &0&-1 &3.677 &0 &0.048 &2.017 &3.218 \\
                                \hline
\end{tabular}}

\end{table*}

\begin{table*}[t]
\captionsetup{skip=2pt} 
\centering
\renewcommand{\arraystretch}{1} 
\caption{Result of dataset size ablation study}
\label{tab:ablation}
\resizebox{0.8\textwidth}{!}{%
\begin{tabular}{llcccccccc}\\ \hline
\textbf{Climate-Zoo Model} & \textbf{ Dataset}          & \textbf{F1 Score \uparrowcell} & \textbf{\begin{tabular}[c]{@{}c@{}}Element \\ Match Score \uparrowcell\end{tabular}}&  \textbf{\begin{tabular}[c]{@{}c@{}}Haversine \\ Distance \(10^3\) \downarrowcell\end{tabular}} & \textbf{BLEU \uparrowcell} & \textbf{ROUGE \uparrowcell} & \textbf{\begin{tabular}[c]{@{}c@{}}GPT4o Score  \\ (Similarity) \uparrowcell\end{tabular}} & \textbf{\begin{tabular}[c]{@{}c@{}}GPT4o Score \\ (Total) \uparrowcell\end{tabular}}  \\ \hline
\multirow{4}{*}{Yi-VL-6B}     & \textit{ClimateIQA}-10k &0.909 &-0.092 &\textbf{1.930} &\textbf{0.820} &\textbf{0.732} &\textbf{4.873} &\textbf{4.685}       \\
                                &  \textit{ClimateIQA}-50k&0.905&-0.070&1.943&\textbf{0.820}&0.728&4.621&4.516         \\
                                &  \textit{ClimateIQA}-100k&\textbf{0.912}&\textbf{-0.048}&1.932&0.814&0.718&4.826&4.377   \\
                                &  \textit{ClimateIQA}-142k&\textbf{0.912} &-0.122 &1.933 &0.815 &0.728 &4.776 &4.348
                                \\\hline
\multirow{4}{*}{LLaVA-v1.6-mistral-7b}     & \textit{ClimateIQA}-10k &0.820    &-0.913    &6.335   &0.611  &0.624      &4.682       &  \textbf{4.597}        \\
                                &  \textit{ClimateIQA}-50k&0.825    &-0.903    &1.945     &0.820  &0.748    & 4.769       & 4.521         \\
                                &  \textit{ClimateIQA}-100k&0.820      &-0.532  &1.972     &\textbf{0.825}      &\textbf{0.750}      &4.648  & 4.401   \\
                                &  \textit{ClimateIQA}-142k&\textbf{0.897}      &\textbf{-0.483}  &\textbf{1.935 }    &0.823 &0.747 &\textbf{4.824} &4.511
                                \\\hline
    
    \end{tabular}}

\end{table*}

\begin{table*}[t]
\captionsetup{skip=2pt} 
\centering
\renewcommand{\arraystretch}{1} 
\caption{Result of question type ablation study using Yi-VL-6B}
\label{tab:ablation_question_type}
\resizebox{0.8\textwidth}{!}{%
\begin{tabular}{lccccccc}\\ \hline
\textbf{ Dataset} & \textbf{F1 Score \uparrowcell} & \textbf{\begin{tabular}[c]{@{}c@{}}Element \\ Match Score \uparrowcell\end{tabular}}&  \textbf{\begin{tabular}[c]{@{}c@{}}Haversine \\ Distance \(10^3\) \downarrowcell\end{tabular}} & \textbf{BLEU \uparrowcell} & \textbf{ROUGE \uparrowcell} & \textbf{\begin{tabular}[c]{@{}c@{}}GPT4o Score  \\ (Similarity) \uparrowcell\end{tabular}} & \textbf{\begin{tabular}[c]{@{}c@{}}GPT4o Score \\ (Total) \uparrowcell\end{tabular}}  \\ \hline
w/o Verification &0.821 &-0.762 &2.124 &0.313 &0.540 &3.268 &3.646       \\
w/o Enumeration &0.892 &-0.984 &2.467 &0.629 &0.576 &2.918 &3.162         \\
w/o Geo-Indexing&0.892&-0.674 &2.676 &0.672 &0.581 &3.186 &3.500   \\
w/o Description &0.889 &-0.866 &1.907 &0.006 &0.002 &1.000 &1.523
\\\hline

    \end{tabular}}

\end{table*}
\noindent\textbf{Supervised Fine-tuning.\quad}Table \ref{tab:sft} illustrates the outcomes of our experiments, highlighting that Climate-Zoo models outperform all baseline models across various metrics. Specifically, for verification and enumeration questions, the baseline models were unable to provide answers, which is reflected in F1 scores of 0 and match scores of -1. In stark contrast, Climate-Zoo models demonstrated an impressive accuracy of around 90\% in pinpointing regions with anomalies, with the highest element match score reaching -0.012, indicating minimal inaccuracies in the data provided. Nevertheless, Climate-Zoo models yield slightly incomplete lists of affected areas. 

In tasks like geo-indexing and description questions, where baseline models did manage to generate responses, they were often plagued by significant errors. On the other hand, Climate-Zoo models significantly outperformed these baseline counterparts by delivering more precise coordinates and more accurate, rich descriptions, achieving superior BLEU, ROUGE, and GPT-4o scores.

While LoRA fine-tuning generally reduces the need for computational resources and, in specific cases like geo-indexing, even outperforms full parameter tuning, it doesn't universally enhance performance across all models. Notably, the Yi-VL-6B LoRA model falls short in handling description questions, underperforming both the fully fine-tuned models and the baseline.

Within the diverse ensemble of the Climate-Zoo collection, each model demonstrates particular strengths. The Qwen-VL-Chat model shines in detecting anomalies within a heatmap and providing detailed, vibrant image narratives, achieving high GPT scores. Conversely, the Yi-VL-6B model stands out with the highest F1 score, showcasing its accuracy in confirming anomalies at pinpoint locations. Meanwhile, the LLaVA-v1.6-mistral-7b model excels in spatial accuracy and textual richness, as evidenced by its exceptional performance in Haversine Distance, BLEU, and ROUGE, making it adept at generating precise coordinates and detailed descriptions. 

\noindent\textbf{Dataset Size Ablation Study.\quad}
Table \ref{tab:ablation} presents the results of an ablation study using models like LLaVA-v1.6-mistral-7b and Yi-VL-6B with full parameters. This study evaluates model performance across varying dataset sizes: 10k, 50k, 100k, and 142k samples. Our findings reveal that increased data volume does not always correlate with improved model performance, with variations observed both between models and across different question types. At the model level, the Yi-VL-6B model achieves excellent results with just 10k samples; increasing the dataset size beyond this point can actually degrade its performance. In contrast, the LLaVA-v1.6-mistral-7b model shows improved performance with larger datasets. At the question type level, verification and enumeration questions demonstrate better performance with larger training datasets, whereas geo-indexing and description questions exhibit more variability. 

Overall, the impact of dataset size on model performance varies significantly among different models. The Yi-VL-6B model appears especially suitable for industrial applications, as it can achieve high effectiveness with smaller datasets and fewer computational resources. We have delved into the potential reasons behind the exceptional performance of the Yi-VL-6B model with the smallest dataset. Our hypothesis centers on the unique characteristics of the pre-training dataset used for Yi-VL-6B. Unlike other VLMs, the Yi-VL-6B model was pre-trained on an extensive dataset comprising 34 billion tokens sourced from encyclopedic texts, which inherently include a significant amount of meteorological and geographical content. This pre-training on domain-rich data likely endowed the model with a robust foundation in meteorological concepts and terminology. As a result, Yi-VL-6B is primed to assimilate new information in this domain with minimal fine-tuning, allowing it to achieve outstanding performance even with a limited dataset.

\noindent \textbf{Question Type Ablation Study.\quad} Table \ref{tab:ablation_question_type} presents the results of our ablation study on various question types. Our findings reveal that all four question types in the dataset are interdependent, and omitting any one of them adversely affects the fine-tuning performance of the VLMs, which subsequently impacts the performance on other questions. Specifically, we observe that excluding any question type significantly impacts the description question, which involve the overall anomalies analysis of the heatmap. The absence of verification questions is particularly detrimental, as it greatly impairs the model's ability to accurately describe anomalies in the heatmaps during the fine-tuning phase.

% 同时我们并没有针对每一个颜色构建相关的地理数据，而是将数据分成了四类。
\section{Conclusions}

In this work, we address the critical challenges posed by meteorological heatmaps in Vision-Language Models (VLMs) by introducing SPOT, a novel algorithm for high-fidelity spatial and structural representation, and \textit{ClimateIQA}, a comprehensive domain-specific VQA dataset. Through SPOT’s sparse localization of irregular features and \textit{ClimateIQA}’s richly annotated tasks, we enable fine-grained reasoning over chaotic visual data. Building on these, our Climate-Zoo suite of fine-tuned VLMs achieves SOTA performance in interpreting meteorological data. This research highlights the transformative potential of domain-specific datasets and innovative architectures in advancing predictive accuracy and actionable insights in meteorology, disaster mitigation, and climate analytics.

\noindent\textbf{Limitations.\quad}
Despite strong performance (91\% accuracy) of our Climate-Zoo models on ClimateIQA, several limitations remain. First, the models struggle with precise color identification in heatmaps, likely because training data only included complete heatmaps. To address this, we propose splitting heatmaps into sub-images for fine-tuning, which may improve color localization.

Additionally, the SPOT algorithm relies on empirically chosen parameters for color segmentation and point selection. Its accuracy depends on initial segmentation, and its advantage over simpler heuristics (e.g., centroids, bounding boxes) requires further quantitative comparison. 
%A thorough sensitivity analysis is also needed.

Meanwhile, our use of RGB heatmaps leverages existing VLMs but introduces issues like projection distortion and dependence on specific color scales, which may limit generalization to other visualization schemes or raw data. Future work should consider training with underlying numerical data.

Moreover, ClimateIQA’s instruction-tuning data is template-based, limiting linguistic diversity and risking overfitting to templates. Broader template variety and testing on paraphrased or out-of-distribution queries are important next steps.

Finally, our dataset currently covers only wind gusts, precipitation, and temperature. Expanding to other phenomena (e.g., drought, typhoons) and integrating VLMs with traditional methods could further enhance robustness and generalization.

Future directions also include integrating \textit{ClimateIQA} with multi-modal data sources such as satellite imagery and ground sensors to address broader ecological challenges (e.g., wildfire risk assessment). Additionally, we plan to develop low-latency (sub-500ms) pipelines and lightweight model variants for real-time disaster response.
\bibliographystyle{ACM-Reference-Format}
\bibliography{sample-base}

\appendix

% In recent years, AI technology has been widely applied in the field of meteorology, such as using AI to predict long-term weather \cite{lam2022graphcast}, forecast typhoon trajectories \cite{bi2022pangu}, or weather classification \cite{dalal2023optimized}. With the development of large language models (LLMs), many meteorology-related LLMs have emerged, including ClimSight \cite{koldunov2024local}, ChatClimate \cite{vaghefi2023chatclimate}, Arabic Mini-ClimateGPT \cite{mullappilly2023arabic}, and ClimateGPT \cite{thulke2024climategpt}. These models learn general knowledge of meteorology and can help answer common questions or provide insights. However, the data used by these models is not updated in real-time. When faced with more challenging problems, such as analyzing the distribution of extreme weather events, relying solely on text data is insufficient to provide accurate answers and can lead to serious hallucinations \cite{bulian2023assessing}. Meteorologists need to analyze extreme weather based on satellite images \cite{liu2024study}, radar data \cite{guastavino2022prediction}, heatmaps \cite{lee2024enhancing}, and isobaric maps, among which images play a crucial role. Nevertheless, there is currently no VLM in the AI field specifically designed to assist meteorologists in interpreting meteorological image like heatmap or help people in related fields analyze weather patterns.

\section{Appendix}

\subsection{Liscense}
The \textit{ClimateIQA}~dataset and Climate-Zoo will be publicly available and use the \textbf{CC BY 4.0 license}.

The IHO Sea Area and the World Bank-approved Administrative Boundaries datasets, used to create \textit{ClimateIQA}, are licensed under \textbf{CC BY 4.0 license}.

The ERA5 dataset is available under a free, worldwide, non-exclusive, royalty-free, and perpetual license. According to this license, access to Copernicus Products is granted for any lawful purpose. Permissible uses include, but are not limited to, reproduction, distribution, public communication, adaptation, modification, and combination with other data and information.

\subsection{Accessibility}

1. Links to access the dataset and its metadata. (\url{https://github.com/AlexJJJChen/Climate-Zoo})

2. The data is saved in both json and csv format, where an example is shown in the README.md file.

3. Precondition Lab research group will maintain this dataset on the official Github account.

4. CC-BY-4.0 (\url{https://github.com/AlexJJJChen/Climate-Zoo/blob/main/LICENSE}).

\subsection{Data Usage}
The authors bear all responsibility in case of violation of rights.

\subsection{Acknowledgements}
We would like to express our sincere gratitude to Jian Chen for his outstanding coordination of this research project, as well as his substantial contributions to manuscript writing, experimental design, and dataset construction. We thank Peilin Zhou and Yining Hua for their dedicated efforts in manuscript preparation, and Dading Chong for conducting the experiments. Special thanks go to Meng Cao for her valuable guidance in computer vision, and to Dr. Yaowei Li for sharing his expertise in meteorology. We also appreciate Wei Chen for his insightful suggestions that helped optimize the manuscript.

We gratefully acknowledge Dr. Bing Zhu, Dr. Junwei Liang, and Dr. Zixuan Yuan for their generous support in providing resources for this project. Dr. Junwei Liang and Dr. Zixuan Yuan serve as the corresponding authors of this paper.

This work was supported by the Guangzhou-HKUST(GZ) Joint Funding Program (Grant Nos. 2024A03J0630 and 2023A03J0008), the National Natural Science Foundation of China 
(Grant No. 62402413) , and the Education Bureau of Guangzhou Municipality. Additional funding was provided by HSBC.

\subsection{Ethics Statements}
When constructing our dataset, we diligently ensure that all data are acquired through legal and ethical means. Committed to the principles of Fair Use, we utilize the dataset strictly for academic research purposes, explicitly prohibiting any form of commercial exploitation.

We acknowledge the responsibility of openly sharing our interface, dataset, codes, and trained models with the public. However, there remains the inherent risk of these resources being misused maliciously. For example, our models could be leveraged to generate responses without appropriately crediting the original information sources. We are dedicated to promoting their ethical use and safeguarding against any harmful or unethical exploitation.

As we progress in the development and application of Vision Language Models (VLMs) within the meteorology domain, it is imperative to address potential ethical concerns to ensure responsible deployment and beneficial outcomes. The misuse of these technologies could propagate disinformation, while inadequate auditing might result in unfair decisions adversely affecting specific groups. Therefore, it is crucial to maintain vigilance in mitigating these issues to uphold ethical standards and equity.

\begin{figure}[t]
    \centering
    \includegraphics[width=1\linewidth]{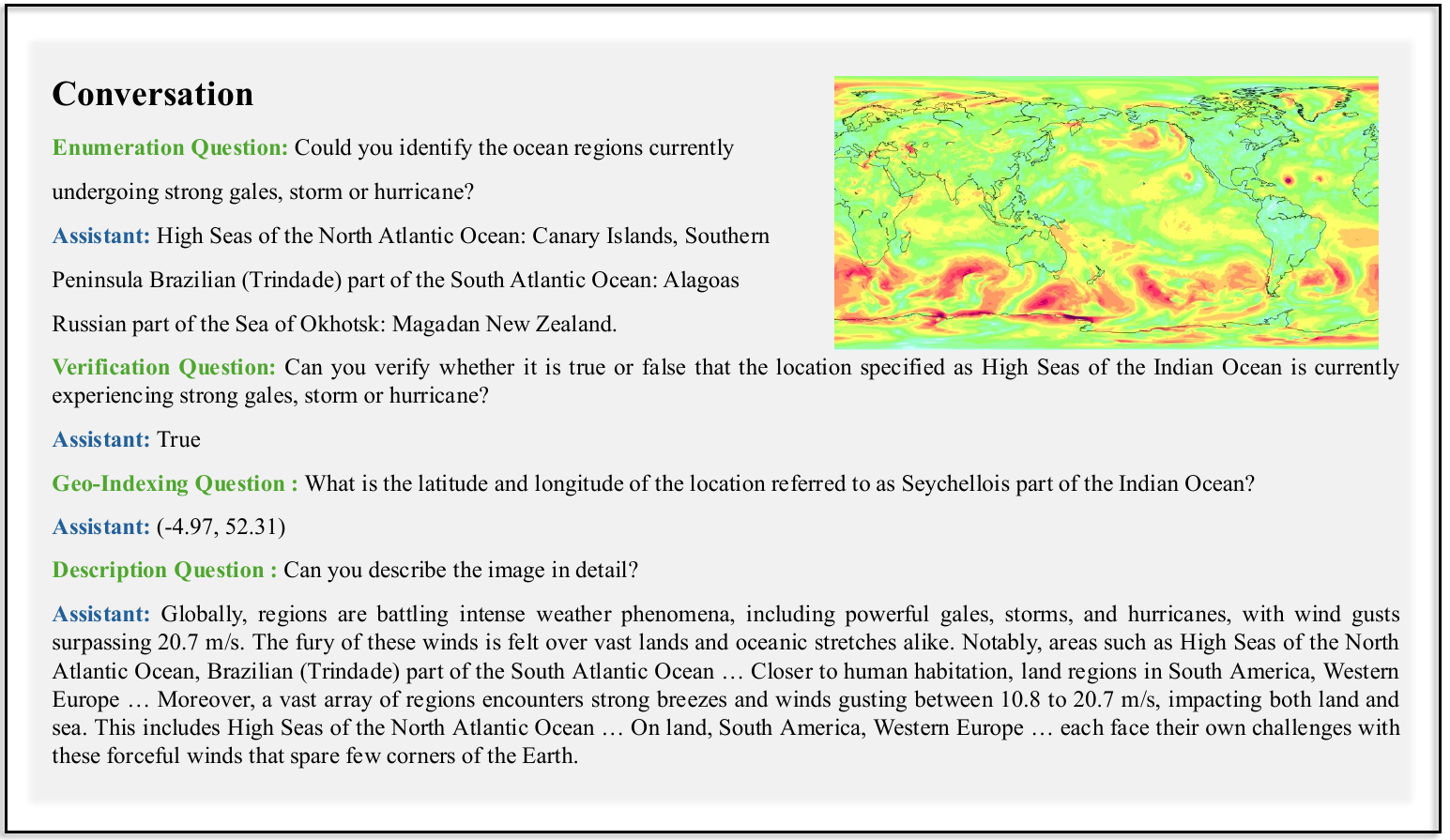}
    \caption{Example of Visual Question Answering Question Types in the Climate Domain.}
    \label{fig:conversation}
\end{figure}

\newpage
\clearpage
\section{Pseudo code of SPOT}

\begin{algorithm}[h]

\caption{SPOT: Sparse Position and Outline Tracking}
\label{alg:spot} % Label the algorithm
\begin{algorithmic}[1] % The number tells where the line numbering should start

\Input Image path $img\_path$, color name $color\_name$
\Output List of latitude and longitude coordinates

\State Initialize $GetCoordinate$ with $img\_path$
\Procedure{GetColorBoundaries}{$image, color\_name$}
    \State Convert $image$ to HSV color space
    \State Generate mask based on color range for $color\_name$
    \State Find contours in the mask
    \State \Return contours, mask
\EndProcedure

\Procedure{GetRepresentativePoints}{$image, contour, num\_points$}
    \State Draw $contour$ on a mask
    \State Erode the mask
    \State Find points in the eroded mask
    \If{number of points $\leq$ $num\_points$}
        \State \Return points
    \Else 
        \State Apply K-Means clustering to points to get $num\_points$
        \State \Return cluster centers as representative points
    \EndIf
\EndProcedure

\Procedure{Process}{$color\_name$}
    \State $contours, mask \gets$ \Call{GetColorBoundaries}{$image, color\_name$}
    \State Calculate total area of selected regions in $mask$
    \For{each $contour$ in $contours$}
        \State Calculate $area\_ratio$ for the contour
        \State Determine $num\_points$ based on $area\_ratio$
        \State $contour\_points \gets$ \Call{GetRepresentativePoints}{$image, contour, num\_points$}
        \State Annotate $image$ with $contour\_points$
    \EndFor
    \State \Return points
\EndProcedure

\Procedure{ConvertPointsToCoordinates}{$points$}
    \State Initialize lists for longitude $\lambda$ and latitude $\varphi$
    \For{each point $pt$ in $points$}
        \State Calculate longitude and latitude based on $pt$ and image dimensions
        \State Append to $\lambda$ and $\varphi$ lists
    \EndFor
    \State \Return $\varphi$, $\lambda$
\EndProcedure

\Procedure{GetCor}{$color\_name$}
    \State $points \gets$ \Call{Process}{$color\_name$}
    \State $\varphi, \lambda \gets$ \Call{ConvertPointsToCoordinates}{$points$}
    \State Print image dimensions
    \State \Return $\varphi$, $\lambda$
\EndProcedure

\end{algorithmic}
\end{algorithm}

\end{document}